# REVIEW OF FACE DETECTION SYSTEMS BASED ARTIFICIAL NEURAL NETWORKS ALGORITHMS

Omaima N. A. AL-Allaf

Assistant Professor,
Faculty of Sciences & IT,
Al-Zaytoonah University of Jordan, P.O. Box130, Amman, (11733), Jordan

## ABSTRACT

*Face detection is one of the most relevant applications of image processing and biometric systems. Artificial neural networks (ANN) have been used in the field of image processing and pattern recognition. There is lack of literature surveys which give overview about the studies and researches related to the using of ANN in face detection. Therefore, this research includes a general review of face detection studies and systems which based on different ANN approaches and algorithms. The strengths and limitations of these literature studies and systems were included also.*

## 1. INTRODUCTION

In the past few years, face recognition has received a significant attention and regarded as one of the most successful applications in the field of image analysis [1]. The human faces represent complex, multidimensional, meaningful visual stimulant. Developing a computational model for face recognition is difficult [2]. Face detection can be regarded as fundamental part of face recognition systems according to its ability to focus computational resources on the part of an image containing a face. The process of face detection in images is complex because of variability present across human faces such as: pose; expression; position and orientation; skin color; presence of glasses or facial hair; differences in camera gain; lighting conditions; and image resolution [3]. The analysis of facial expression was primarily a research field for psychologists in the past years [4]. At the same time, advances in many domains such as: face detection [5][6]; tracking[7]; and recognition [1]; pattern recognition; and image processing contributed significantly to research in automatic facial expression recognition.

Face detection should be performed before recognition system. This is done to extract relevant information for face and facial expression analysis. Two classes of techniques for face representation and relevant information extraction. And geometrical feature extraction relies on parameters of distinctive features such as eyes, mouth and nose. At the same time, a face is represented as an array of pixel intensity values suitably pre-processed in appearance based approaches (texture). This array is then compared with a face template using a suitable metric [4]. Research [8] compared the performances of these representation techniques in face recognition. Therefore, according to the complexity of face detection process, many applications based on human face detection have been developed recently such as: surveillance systems, digital monitoring, intelligent robots, notebook, PC cameras, digital cameras and 3G cell phones. These applications play an important role in our life. Nevertheless, the algorithms of the applications are complicated and hard to meet real-time requirements of specific frame-rate [9]. Over the past decade, many approaches for improving the performance of face detection have been proposed [9][10][11][12][13][14][15][16][17][18][19]. At the same time, many literature studies focused on survey on face detection techniques [20][21][6][22][23].

Artificial neural networks (ANN) were used largely in the recent years in the fields of image processing (compression, recognition and encryption) and pattern recognition. Many literature researches used different ANN architecture and models for face detection and recognition to achieve better compression performance according to: compression ratio (CR); reconstructed image quality such as Peak Signal to Noise Ratio (PSNR); and mean square error (MSE). Few literature surveys that give overview about researches related to face detection based on ANN. Therefore, this research includes survey of literature studies related to face detection systems and approaches which were based on ANN.

The rest of this paper is organized as follows: Section 2 includes the main steps of face detection and recognition. Section 3 includes literature studies related to face detection systems based on ANN. Section 4 includes comparisons between these literature studies. Section 5 includes recommendations. Finally section 6 concludes this work.

## 2. FACE DETECTION AND RECOGNITION

A general face recognition system includes many steps: face detection; feature extraction; and face recognition [14][24] as shown in Figure1. Face detection and recognition includes many complementary parts, each part is a complement to the other. Depending on regular system each part can work individually. Face detection is a computer technology that is based on learning algorithms to allocate human faces in digital images [25].

Face detection takes images/video sequences as input and locates face areas within these images. This is done by separating face areas from non-face background regions. Facial feature extraction locates important feature (eyes, mouth, nose and eye-brows) positions within a detected face. Feature extraction simplifies face region normalization where detected face aligned to coordinate framework to reduce the large variances introduced by different face scales and poses. The accurate locations of feature points sampling the shape of facial features provide input parameters for the face identification. Other face analysis task: facial expression analysis [16]; face animation and face synthesis can be simplified by accurate localization of facial features [16][24].

Face identification generates the final output of complete face-recognition system: the identity of the given face image. Based on normalized face image and facial feature locations derived from previous stages, a feature vector is generated from given face and compared with a database of known faces. If a close match is found, the algorithm returns the associated identity. A main problem in face identification is the large differences between face images from the same person as compared to those from different persons. Therefore, it is important to choose a suitable face classification technique that can provide a good separate ability between different persons. Face identification has a wide range of applications. Because it offers a non-intrusive way for human identification, the face is used as an important biometric in security applications.

Recently, face recognition has received wide interest in number of countries are integrating facial information into electronic passport and to other biometrics (fingerprints and iris) [26]. In addition to security and law enforcement, face recognition is also applied in entertainment and consumer electronics as a means for a natural user interface. By recognizing the existence of the user and his identity, consumer devices can offer customized services, thereby creating an enhanced user experience. To achieve high-performance face recognition system, each processing stage in the system has to be designed to satisfy application requirements [14]. Face recognition involves comparing an image with a database of stored faces to identify individual in input image. The related task of face detection has direct relevance to face recognition because images must be analysed and faces identified, before they can be recognized. Detecting faces in an image can help to focus the computational resources of the face recognition system, optimizing the systems speed and performance [3].

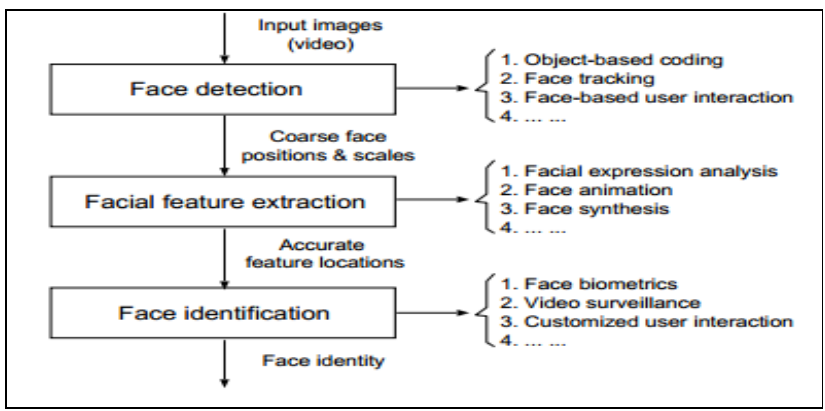

Figure 1. Framework of a face-recognition system [14]

The main steps of face detection system are shown in Figure2. Face detection separate image windows into two parts: one containing faces, and one containing the background. The process is difficult because the: commonalities exist between faces (vary in terms of age, skin color and facial expression); and also differing in: lighting conditions; image qualities; and geometries. The face detector would be able to detect the presence of any face under any set of lighting conditions, upon any background.

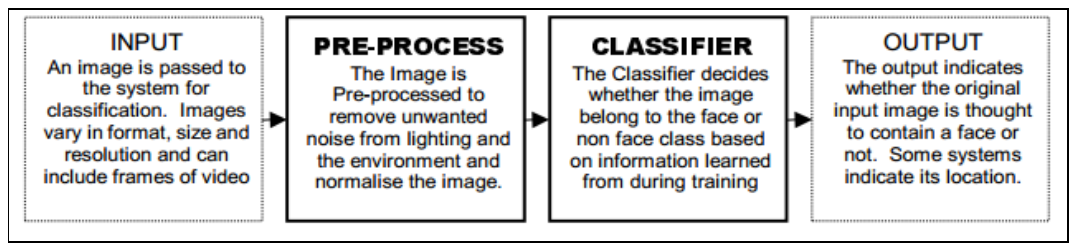

Figure 2. A general face detection system [3]

## 3. ARTIFICIAL NEURAL NETWORKS FOR FACE DETECTION

In the recent years, different architectures and models of ANN were used for face detection and recognition. ANN can be used in face detection and recognition because these models can simulate the way neurons work in the human brain. This is the main reason for its role in face recognition. This research includes summery review of the researches related to face detection based on ANN.

### 3.1. Retinal Connected Neural Network (RCNN)

Rowley, Baluja and Kanade (1996) [27] presented face detection system based on a retinal connected neural network (RCNN) that examine small windows of an image to decide whether each window contains a face. Figure 3 shows this approach. The system arbitrates between many networks to improve performance over one network. They used a bootstrap algorithm as training progresses for training networks to add false detections into the training set. This eliminates the difficult task of manually selecting non-face training examples, which must be chosen to span the entire space of non-face images. First, a pre-processing step, adapted from [28], is applied to a window of the image. The window is then passed through a neural network, which decides whether the window contains a face. They used three training sets of images. Test SetA collected at CMU: consists of 42 scanned photographs, newspaper pictures, images collected from WWW, and TV pictures (169 frontal views of faces, and require ANN to examine 22,053,124 20×20 pixel windows). Test SetB consists of 23 images containing 155 faces (9,678,084 windows). Test SetC is similar to Test SetA, but contains images with more complex backgrounds and without any faces to measure the false detection rate: contains 65 images, 183 faces, and 51,368,003 window. The detection ratio of this approach equal 79.6% of faces over two large test sets and small number of false positives.

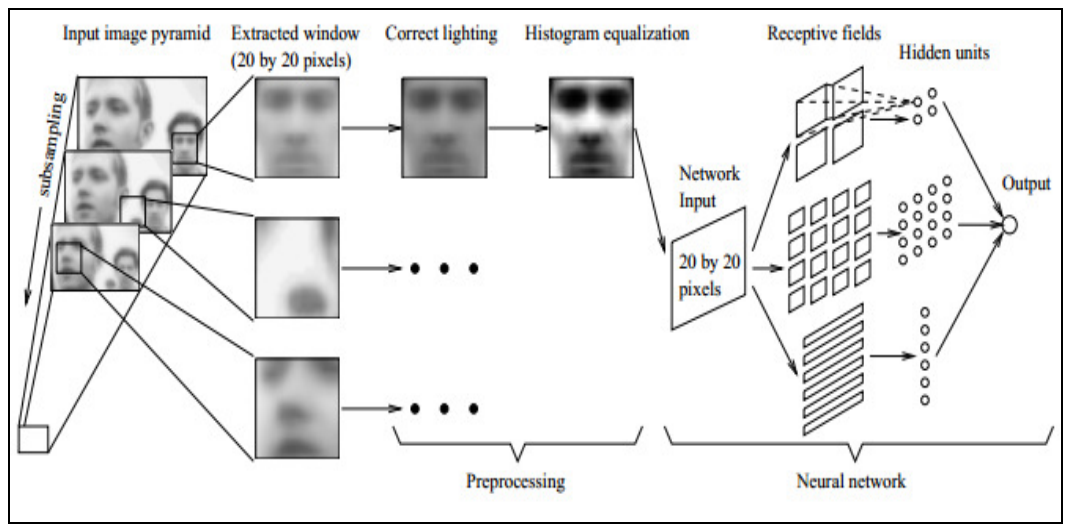

Figure 3. RCNN for face detection [27]

### 3.2. Rotation Invariant Neural Network (RINN)

Rowley, Baluja and Kanade (1997) [29] presented a neural network-based face detection system. Unlike similar systems which are limited to detecting upright, frontal faces, this system detects faces at any degree of rotation in the image plane. Figure 4 shows the RINN approach. The system employs multiple networks; the first is a "router" network which processes each input window to determine its orientation and then uses this information to prepare the window for one or more detector networks. We present the training methods for both types of networks. We also perform sensitivity analysis on the networks, and present empirical results on a large test set. Finally, we present preliminary results for detecting faces which are rotated out of the image plane, such as profiles and semi-profiles [29].

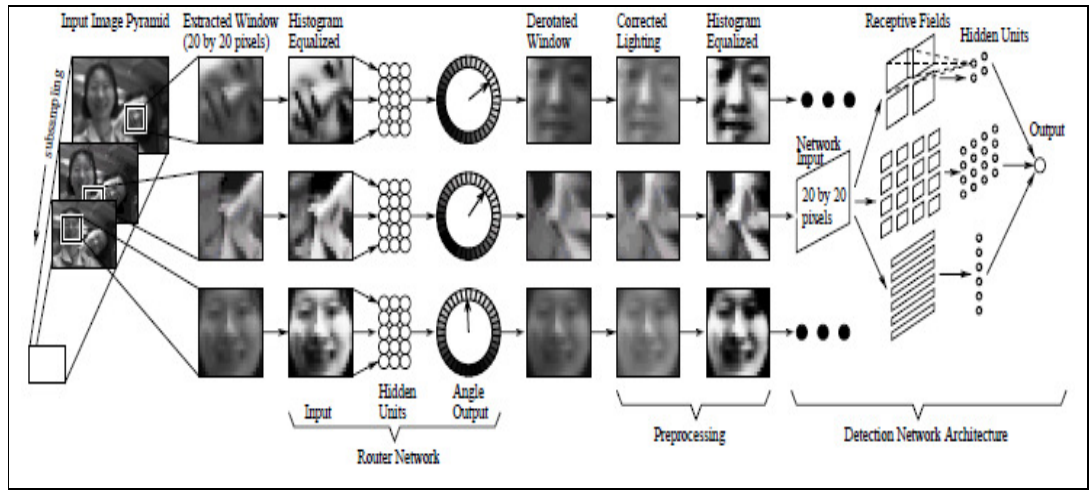

Figure 4. RINN for face detection [29]

## 3.3. Principal Component Analysis with ANN (PCA & ANN)

Jeffrey Norris (1999) [30] used using principal component analysis (PCA) with class specific linear projection to detect and recognized faces in a real-time video stream. Figure 5 shows PCA & ANN for face detection. The system sends commands to an automatic sliding door, speech synthesizer, and touchscreen through a multi-client door control server. Matlab, C, and Java were used for developing system. The system steps to search for a face in an image:

1. Select every 20×20 region of input image.
2. Use intensity values of its pixels as 400 inputs to ANN.
3. Feed values forward through ANN, and
4. If the value is above 0.5, the region represents a face.
5. Repeat steps (1..4) several times, each time on a resized version of the original input image to search for faces at different scales [30].

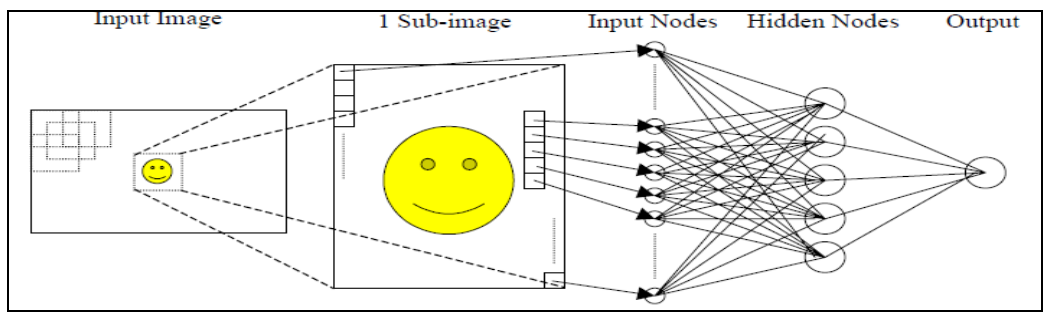

Figure 5. PCA & ANN for face detection [30]

## 3.4. Fast Neural Networks (FNN)

Hazem El-Bakry (2002) [31] proposed fast neural networks (FNN) approach to reduce the computation time for locating human faces. Each image is divided into small sub images and then each one is tested separately using a fast ANN. The experimental results of comparison with conventional neural networks showed that the high speed is achieved when applying FNN.

### 3.5. Polynomial Neural Network (PNN)

Lin-Lin Huang, et al. (2003) [32] proposed face detection method using a polynomial neural network (PNN). The local regions in multi scale sliding windows are classified by the PNN to two classes (face and non-face) to locate the human faces in an image. The PNN takes the binomials of projection of local image onto a feature subspace learned by principal component analysis (PCA) as inputs. They investigated the influence of PCA on either the face samples or the pooled face and non-face samples.

### 3.6. Convolutional Neural Network (CNN)

Masakazu Matsugu (2003) [33] described a rule-based algorithm for robust facial expression recognition combined with face detection using a convolutional neural network (CNN). Figure 6 shows the CNN approach. The problem of subject independence and translation, rotation, and scale invariance in facial expression recognition were addressed in this study.

### 3.7. Evolutionary Optimization of Neural Networks

Stefan, et al (2004) [34] used ANN to get decision whether a pre-processed image region represents a human face or not. They described the optimization of this network by a hybrid algorithm combining evolutionary computation and gradient-based learning. The evolved solutions perform considerably faster than an expert-designed architecture without loss of accuracy. The proposed hybrid algorithm tackles the problem of reducing the number of hidden neurons of face detection network without loss of detection accuracy. The speed of classification whether an image region corresponds to a face or not could be improved by approximately 30 %. Figure 7 shows evolutionary optimization of ANN.

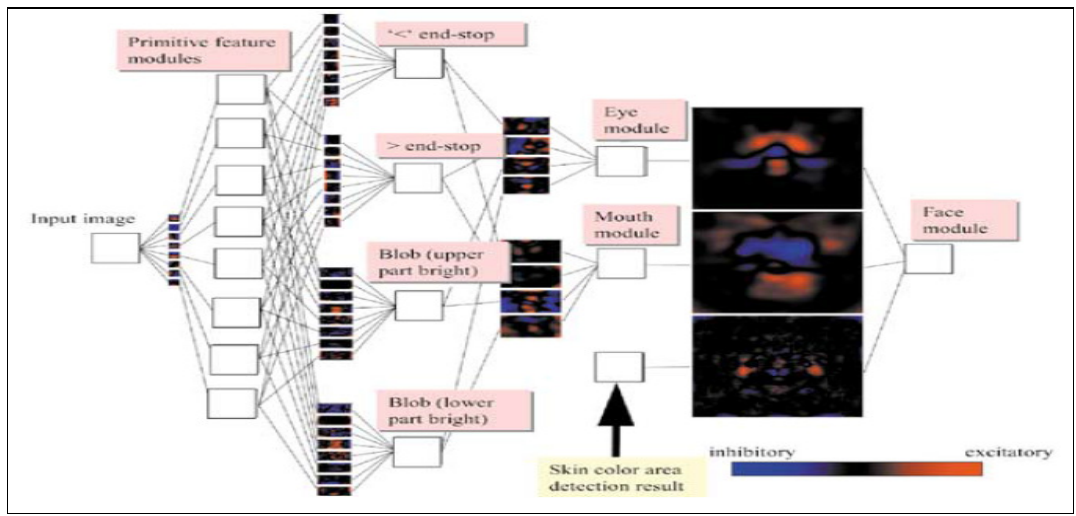

Figure 6. CNN  for face detection [33]

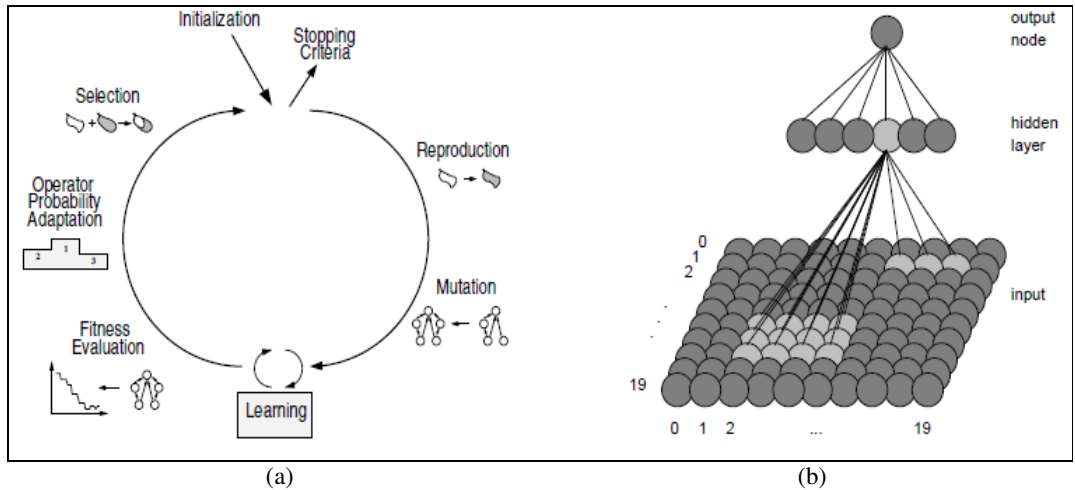

Figure 7. (a): Hybrid algorithm. (b): Visualization of input and field connectivity [34]

.

### 3.8. Multilayer Perceptron (MLP)

According to Rowley work in [27], Marian Beszedes & Milos Oravec (2005) [35] presented a neural network based face detection system to detect faces in an unprocessed input image. Figure 8 shows the MLP approach. They used image processing techniques such as normalization, rotation and position, light conditions improvement on small windows extracted from the input image. Multilayer perceptron (MLP) used to detect rotation of input window and also to decide whether the window contains a face or not. This system is based on method to distribute the decision among multiple sub networks and an algorithm is used to train this ANN and the result of this system is in the form of a set containing locations of human faces. The training/testing set for includes 250 face and 250 randomly generated non-face samples. The training/testing sets were extended according to boosting method by 2000 and 2500 non- face samples. These samples were obtained as false face detections from testing of ANN trained on basic sets on images containing no faces. Their system succeeded in 90100 percent when they consider the number of faces in input picture.  Only 18 from this amount were not classified correctly. Only three faces from the total number 38 were not found.

Shilbayeh and Al-Qudah (2008)  [36] proposed a face detector based multi layer perceptron (MLP) ANN and maximal rejection classifier (MRC) to improve efficiency of detection in comparison with traditional ANN. The ANN was organized to reject a majority of non-face patterns in image backgrounds by improving detection efficiency, reducing computation cost and maintaining the detection accuracy.

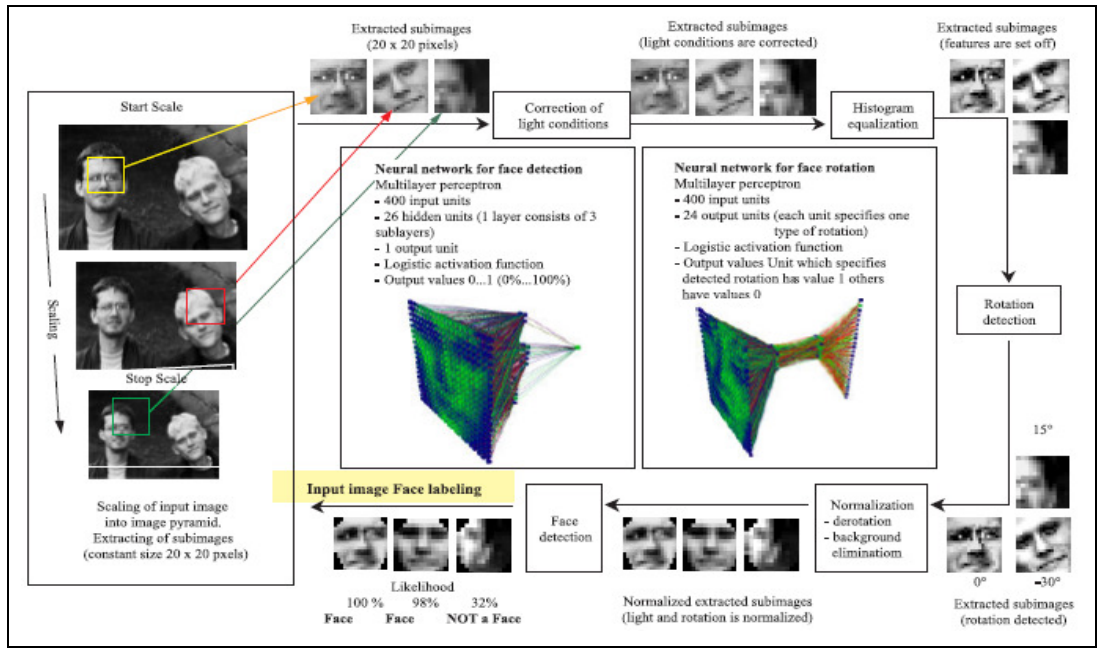

Figure 8. MLP of face localization system [35]

## 3.9. Back Propagation Neural Networks (BPNN)

Zoran and Samcovic (2006) [37] used ANN for face detection for video surveillance. The ANN is trained with multilayer back propagation neural networks (BPNN). Three face representations were taken (pixel, partial profile and eigenfaces) representation. Three independent sub-detectors are generated based on these three face representation. Figure 9 shows the BPNN approach. Aamer Mohamed, et al (2008) [38] proposed face detection system based on BPNN via Gaussian mixture model to segment image based on skin color. In this approach start from skin and non skin face candidates' selection. After that the features are extracted from discrete cosine transform (DCT) coefficients. Based on DCT feature coefficients in Cb and Cr color spaces, BPNN was used to train and classify faces. The BPNN used to check if the image include face or not. DCT feature values of faces that represent the data set of skin/non-skin face candidates obtained from Gaussian mixture model are fed into BPNN to classify whether original image includes a face or not.

## 3.10. Gabor Wavelet Faces with ANN

Sahoolizadeh et al (2008) [39] proposed hybrid approaches for face recognition based on combined Gabor wavelet faces with ANN feature classifier. The Gabor wavelets used to represent face image. The representation of face images using Gabor wavelets is effective for facial action recognition and face identification.

They reduced dimensionality and linear discriminate analysis on down sampled Gabor wavelet faces can increase the discriminate ability. Nearest feature space is extended to various similarity measures shows good performance which achieve 93% recognition rate on ORL data set.

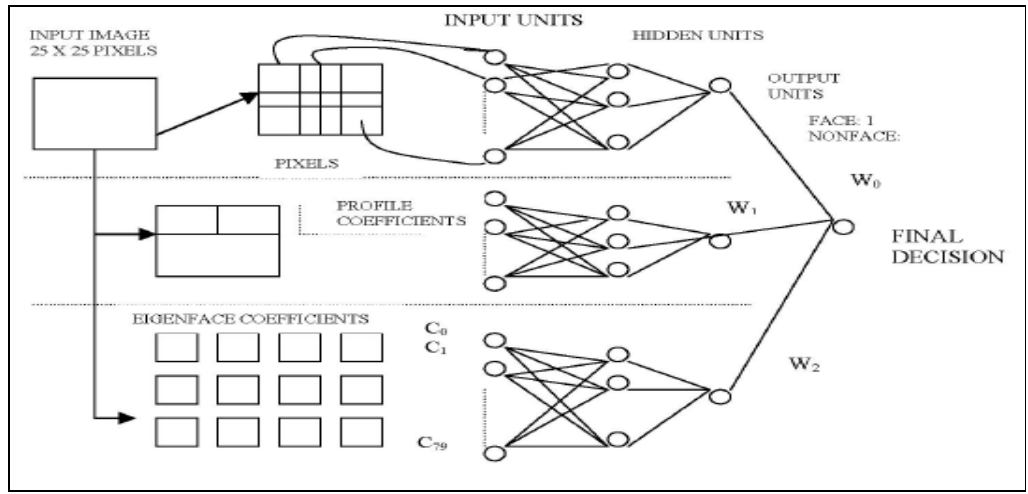

Figure 9. BPNN architecture for face detection [37]

Avinash and Raina (2010) [40] presented face detection approach with Gabor wavelets transform & feed forward neural network for finding feature points and extracting feature vectors. Gabor filter used for feature extraction for face detection. The classifier (FFNN) take the feature vectors as input. The location of feature points contains information about the face in this approach. The graph is constructed from the general face idea. Instead of fitting this graph, the feature points are obtained from the characteristics of each face automatically. Facial features allow to make a decision from face parts because the facial features are compared locally instead of using a general structure. Two measures were used in this study : false negative and false positive. These two measures can be calculated using Equation 1 and Equation 2 as follows [39]:

$$False\ Negative = \frac{Number\ of\ Missed\ Faces}{Total\ Number\ of\ Actual\ Faces} \quad \dots\dots\dots\dots\dots\dots\dots\dots\dots\dots\dots\dots(1)$$

$$False\ Positive = \frac{Number\ of\ Incorrect\ Detected\ Faces}{Total\ Number\ of\ Actual\ Faces} \quad \dots\dots\dots\dots\dots\dots\dots\dots\dots\dots\dots(2)$$

Mohammad Abadi, et al, (2011) [41] proposed approach based on ANN and Gabor wavelets to detect desirable number of faces in fixed photo with gray background. They used correlation of a window with a face with photo. Then they estimated areas of candidate of face presence. After that, they used step algorithm and referred these areas and around them to section of extraction of Gabor wavelets characteristics and neural network classifier. The resultant areas lead to detection of face locations in photo. They examined the result of estimation of efficiency of this method by different tests. The method is simulated in MATLAB. They used 70 face photos and 60 non face photos in training phase. Every face photo, its mirror photo and with the angle of 5,10,15 degrees in positive and negative directions and photos with one pixel shift in every 4 directions are placed in training set for reducing network sensitivity. For no face photos also, their mirror and their 180 degrees transformation is placed in training data. They obtained 5% right answer, error limit of 0.0001, false negative error= 5%. And for tested image with size=254×600, the positive false=12, and detection= 56 from 57. Also for test image of size = 50×100, the positive false=0, and detection= 2 from 3.

Anissa Bouzalmat, et al (2011) [42] presented BPNN for face recognition. The BPNN input is feature vector based on Fourier Gabor filters. They used an algorithm for detecting face regions in images using the color of skin which presents overlooked in different background, accessory and clothing. After that, they introduced Gabor filters with 8 orientations and 5 resolutions to get maximum information (to extract the maximum of information by varying the resolution and orientation). This is done to generate and extract the features vector of the whole face in image. BPNN is then applied to perform the recognition task. This solution was implemented using Java environment. Results indicate that the proposed method achieves good results. Figure 10 shows the BPNN with Gabor Wavelet for face detection.

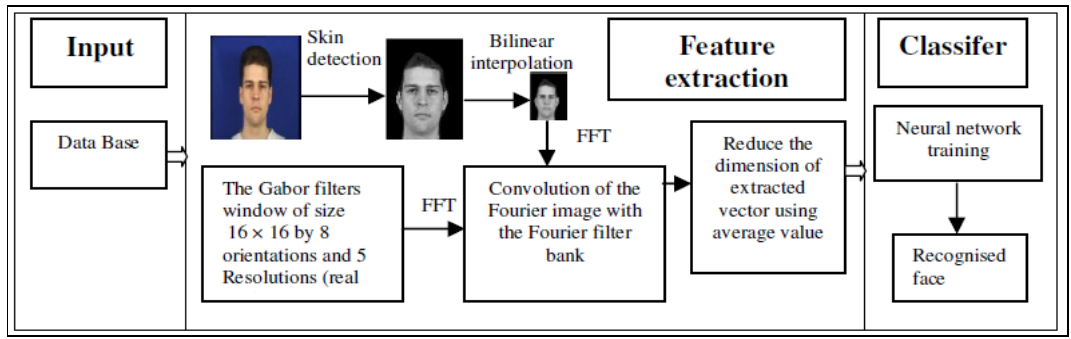

Figure 10. Description of the proposed solution architecture [42]

### 3.11. Skin Color and BPNN

The construction of robust real-time face detection system can be regarded as one of the most practical applications under vigorous development. Kalavdekar Prakash (2010) [43] described face detection system that process images based neural network to detect face images and achieving high detection rates. Analyzing a video sequence is the current challenge since faces are constantly in dynamic motion, presenting many different possible rotational and illumination conditions. While solutions to the task of face detection have been presented, detection performances of many systems are dependent upon environment. Their suggested system includes: skin color filter; image filtering; MLP; and detection. Figure 11 shows Skin Color and BPNN for face detection. Face detection attempt to locate all regions that contain a face in given still image or an image sequence. There are two main solutions for face detection: feature-based and image-based approaches.

Elmansori and Khairuddin (2011) [44] presented face detection method combines two algorithms: Skin color based face detector and BPNN. The Skin color based face detector used modeling the distribution of skin color to identify areas most likely to be regions of skin to identify potential areas of skin by equalizing probability of likelihood. The problem space is linearly separable and a linear threshold function is offered for the solution which is supported by a sparse feature mapping architecture. BPNN is used to represent function using arbitrary decision surfaces by utilizing nonlinear activation functions. Their experiments showed that the methods show closer performances for the classification in face and non-face space, and the method has achieved high detection rates and an acceptable number of false negatives and false positives. The system was implemented using C#.

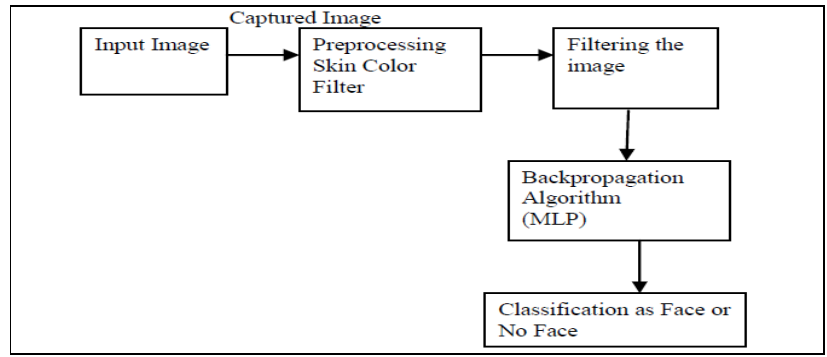

Figure 11. Skin Color and BPNN for Face Detection [43]

### 3.12 Cascaded Neural Network

Finally, Zuo & de With (2008) [45] proposed a fast face detector based on a hierarchical cascade of neural network ensembles to enhance detection accuracy and efficiency. They used a number of neural network classifiers to form a neural network ensemble. Each classifier is specialized in a sub region in the face-pattern space. These classifiers complement each other to perform the detection task. Then, they organized the neural network ensembles in a pruning cascade to reduce the total computation cost of face detection.

In this stage, simpler and more efficient ensembles used at earlier stages in the cascade are able to reject a majority of non face patterns in image backgrounds by improving the overall detection efficiency while maintaining detection accuracy. Their results showed that the proposed neural-network ensembles improve the detection accuracy as compared to traditional ANN. Their approach reduced training and detection cost by achieving detection rate equal 94%.

## 4. COMPARISONS BETWEEN DIFFERENT ANN APPROACHES

Literature studies based on ANN for building face detection systems were described in this study. Each one of these studies was based on special architecture of ANN for face detection. Many of these studies were based on one architecture of ANN such as : Multilayer Perceptron (MLP); BackPropagation Neural Networks (BPNN); Retinal Connected Neural Network (RCNN); Rotation Invariant Neural Network (RINN); Fast Neural Networks (FNN); Convolutional Neural Network; Polynomial Neural Network (PNN). Other studies were based on ANN on combination with other techniques and methods such as Principal Component Analysis with ANN (PCA & ANN); Evolutionary Optimization of Neural Networks; Gabor Wavelet Faces with ANN; and finally Skin Color and BPNN.

All of these studies were based on ANN. Each one of these studies includes its own experiments and based on different database for training and testing images. Many of these studies take detection rate as performance measure; other studies take error rate as performance measure and so on. Other studies did not talk exactly about the used database.

On the other hand, each of these literature studies has its own strengths and limitations. We cannot exactly determine what the best topology that can be used for face detection system with high performance. Table 1 shows information about topology, database of training/testing images and performance of face detection systems which were taken from many of these literature studies.

Table 1 shows that the literature studies used different data bases for image training and testing set. Other studies were not included in table 1 because they weren't use known database and just used camera image samples. The number of samples is different from one study to another.as example the research [38] take only 50 image samples. Whereas other studies such as [27][32] used more than one image sets as samples. We can note from table 1 also that the studies adopted different image dimensions: (384×384), (92×112) and so on.

Table 1.  DB and Performance measures used in literature studies

| Research | Topology | Data Base: Training & Testing | Performance |
|---|---|---|---|
| [27] | Retinal connected neural network | Three training sets of images. Test SetA: 42 scanned photographs Test SetB: 23 images contain 155 faces Test SetC: 65 images, 183 faces  (images with more complex backgrounds and without faces to measure false detection ) | Detect 78.9% - 90.5% of faces in a set of 130 test images<br><br>Acceptable number of false detections. |
| [30] | PCA with ANN | Select 700 pictures in Kah-Kay Sung's data set of 1488 faces to train ANN with 700 random noise pictures as negative examples remaining 788 faces in Kah-Kay's data set,  followed by 788 random noise pictures | 1.2% error after training for 50 epochs<br><br>1566 examples, 35 mis classifications made (2.23% error). |
| [32] | PNN | First set: 3257 images downloaded from several websites (384×384), with one face in each image. Second set: 130 images downloaded from website of CMU | Detection rate = 84.6% False rate=3:51 × 10−6 |
| [33] | CNN | training of CNN, the number of facial fragment images used is 2900 for the FD2 layer, 5290 for the FD3, and 14,700 (face) for the FD4 layer, respectively. Number of non-face images, also used for the FD4 layer, is 137. | Recognition rate = 97.6% for 5600 still images of more than 10 subjects |
| [37] | BPNN | Training set contains 12000 face images collected from various face DBs. These samples also include the scaled versions at the same face with factor (0.8 - 1.12) | Detection rates measured for separate test set of 500 faces and 4000 non-faces. Performance=94%. |
| [39] | Gabor wavelet with ANN | ORL dataset: 400 frontal faces: 10 tightly cropped (92×112) with 256 grey images of 40 individuals with variations in pose, illumination, ..etc | Detect (77.9% - 90.3%) of faces in a set of 130 test images |
| [36] | MLP and MRC | Training set: face images from MIT DB. Images (scaled to 20×20) Test set: 2000 face/non-face images from MIT DB. Non-face patterns generated at different locations and scales. | Detection rate = 91.6%<br><br>Error rate = 7.54% |
| [38] | BPNN | 50 real images taken under different lighting conditions (digital camera images and web images from several websites). | Detect 97.3% of faces in a set of 50 real images. Processing Time (s) of image (63×180)= 3.8 Processing Time (s) of image (200×219)= 6.2 |

Finally we can note from table 1 that the factors used as performance evaluations are different from one study to other. Many studies used detection rate, others used false rate, and so on. Figure 12 shows the detection rate of different ANN approaches from many studies. We can note from Figure 12 that the highest face detection rate can be obtained from using the CNN [33] approach. At the same time the BPNN approach adopted in [38] and [37] result in obtaining good and high detection rate.

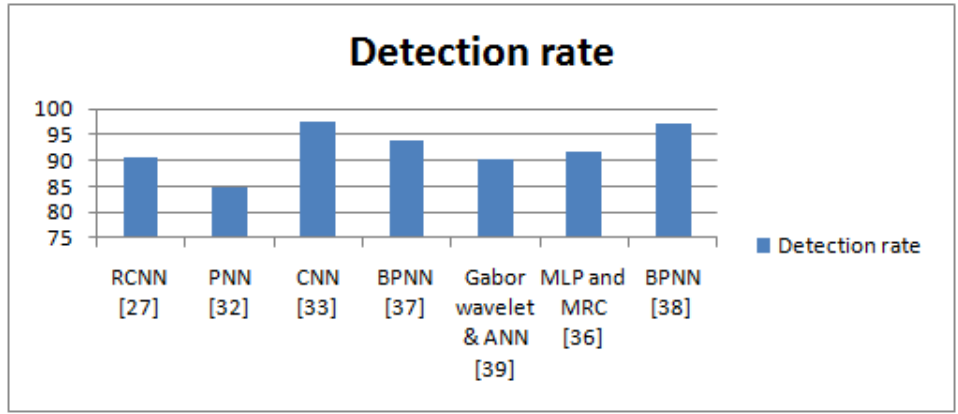

Figure 12. Detection rate for different studies

## 5. RECOMMENDATIONS FOR FACE DETECTION SYSTEM

Face detection is first step in face recognition systems to localize and extract the face region from the image background. The literature studies related to face detection systems which were based on ANN were described earlier in this research can be summarized as follows:

- The face detection techniques presented above were based on 2D image detection.
- Many of these literature researches did not give overview about the exactly used database for system training and testing.
- Many of these researches did not give sufficient information about performance measures used for face detection. There is lack of equations related to these performance measures.
- Most of these studies of face detection systems were adopted ANN in combination with other approaches and algorithms to obtain better results for detection and improve the performance of face detection system. But this may increase the system complexity, required memory and time for face detection.
- Lack of using other significant ANN architectures such as self-organizing map, PatternNet ANN, Fast BPNN and so on.
- Lack of literature related to face detection based on combination of ANN and genetic algorithm (GA).

According to above points, many recommendations must be taken in our consideration when we suggest to build a strong face detection system such as:

- Try to design a real time face detection system that is based on video taken in real time camera.
- Give sufficient details about the exactly used database for system training and testing.
- Give sufficient details about performance measures and equations used for face detection.

- ANN can be adopted in combination with other algorithms to obtain better results for face detection. At the same time, we must focus on how to simplify the combined algorithms steps to reduce the memory required and processing time.
- Try to use other ANN architectures: self-organizing map, PatternNet, FitNet and Fast BPNN.
- Try to use different optimization ANN training algorithms such as: Levenberg-Marquardt (TRAINLM); TRAINBFG; Bayesian regularization (TRAINBR); TRAINCGF algorithm; Gradient descent (TRAINGD); and Gradient descent with momentum (TRAINGDM) [46] to obtain best results for face detection system.
- Try to use genetic algorithm (GA) as an optimization algorithm to obtain the best values of ANN algorithm parameters that result to optimal results.

# 6. CONCLUSION

This paper includes a summary review of literature studies related to face detection systems based on ANNs. Different architecture, approach, programming language, processor and memory requirements, database for training/testing images and performance measure of face detection system were used in each study. Each study has its own strengths and limitations.

In future work, a face detection system will be suggested based on using Pattern Net and Back propagation neural network (BPNN) with many hidden layers. Different network architectures and parameters' values of BPNN and PatternNet will be adopted to determine PatternNet architecture that will result in best performance values of face detection system.

## ACKNOWLEDGEMENTS

The author would like to thanks Al-Zaytoonah university of Jordan- Amman- Jordan for supporting this research.

## REFERENCES

[1]    W. Zhao, et al (2000) "Face recognition: a literature survey", Technical Report CAR-TR-948, University of Maryland, October 2000.

[2]    Turk M & Pentland A (1991) "Eigenfaces for recognition", Journal of Cognitive Neuroscience, Vol.3, pp71–86.

[3]    Phil Brimblecombe (2002) "Face Detection using Neural Networks", H615 – Meng Electronic Engineering, School of Electronics and Physical Sciences, URN: 1046063.

[4]    Bouchra Abboud, et al (2004) "Facial expression recognition and synthesis based on an appearance model", Signal Processing: Image Communication, Vol. 19, Issue. 8, pp723-740.

[5]    P. Viola & M.J. Jones (2001) "Robust real-time object detection", Technical Report CRL/2001/01, Cambridge Research Laboratory, USA, February 2001

[6]    Ming-Hsuan Yan, et al (2002), "Detecting Faces in Images: A Survey", IEEE Transactions on Pattern Analysis and Machine Intelligence, Vol. 24, No. 1, pp34-58, January2002

[7]    Minyoung Kim, et.al (2008) "Face tracking and recognition with visual constraints in real-world videos", IEEE Conference on Computer Vision and Pattern Recognition, pp23-28, June.

[8]    Brunelli R & Poggio T (1993) "Face recognition: features versus templates", IEEE Transaction Pattern Analysis and Machine Intelligence, Vol. 15, No.10, pp1042–1052.

[9]    Yao-Jiunn Chen & Yen-Chun Lin (2007) "Simple Face-detection Algorithm Based on Minimum Facial Features", The 33rd Annual Conference of the IEEE Industrial Electronics Society (IECON) 5-8 Nov 2007, Taipei, Taiwan, pp455-460.

[10]   Sanjay Singh, et. Al (2003) "A Robust Skin Color Based Face Detection Algorithm", Tamkang Journal of Science and Engineering, Vol. 6, No. 4, pp.227-234 .

[11] Abdenour Hadid, Matti Pietik¨ainen & Timo Ahone (2004) "A Discriminative Feature Space for Detecting and Recognizing Face", Proceedings of the 2004 IEEE Computer Society Conference on Computer Vision and Pattern Recognition, Vol. 2.

[12] Elise Arnaud et al (2005) "A Robust And Automatic Face Tracker Dedicated To Broadcast Videos, IEEE International Conference On Image Processing.

[13] Zhonglong Zheng, Jie Yang & Yitan Zhu (2006) "Face Detection and Recognition using Colour Sequential Images", Journal of Research and Practice in Information Technology, Vol. 38, No. 2, pp.135-149, May 2006.

[14] Fei Zuo (2006) Embedded Face Recognition Using Cascaded Structures, Thesis, Technische Universiteit Eindhoven, China.

[15] Zuo F. and P.H.N. de With (2002) "Automatic Human Face Detection for a Distributed Video Security System", Proceedings of the Progress workshop on Embedded Systems, pp269–274, Oct.

[16] Stan Z. Li & Anil K. Jai (2005) Handbook of Face Recognition, Springer Science &Business Media.

[17] Jun-Su Jang & Jong-Hwan Kim (2008) "Fast and Robust Face Detection Using Evolutionary Pruning", IEEE Transactions on Evolutionary Computation, pp1-10.

[18] Bernd Menser & Michael Brunig (2000) "Face Detection and Tracking for Video Coding Application", Conference Record of the Thirty-Fourth Asilomar Conference on Signals, Systems and Computers, 29Oct-1Nov2000, Pacific Grove, CA, USA

[19] Pedro Alexandre Dias Martin (2008) "Active Appearance Models for Facial Expression, Recognition and Monocular Head Pose Estimation", master thesis, Dept. of Electrical and Computer Eng., Faculty of Sciences and Technology, University of Coimbra.

[20] Hjelmas and Low (2001) "Face Detection: A Survey," Computer Vision and Image Understanding, vol. 83, pp236-274, doi:10.1006/cviu.2001.0921, http://www.idealibrary.com

[21] Yongzhong Lu, Jingli Zhou & Shengsheng Yu (2003) "A Survey of Face Detection, Extraction and Recognition", Computing and Informatics, Vol. 22, pp.163-195.

[22] W. Zhao et al (2003) "Face recognition: A literature survey", ACM Computing Surveys, Vol.35, No.4, December 2003, pp. 399–458.

[23] Cha Zhang & Zhengyou Zhang (2010) "A Survey of Recent Advances in Face Detection", Technical Report, MSR-TR-2010-66, Microsoft Research, Microsoft Corporation One Microsoft Way, Redmond, WA 98052, http://www.research.microsoft

[24] Abboud B, Davoine F & Dang M (2004) "Facial expression recognition and synthesis based on an appearance model", Signal Processing: Image Communication, Vol.19, No.8, pp723–740.

[25] Mohammad Alia, Abdelfatah Tamimi and Omaima Al-Allaf, "Integrated System For Monitoring And Recognizing Students During Class Session", AIRCC's: International Journal Of Multimedia & Its Applications (IJMA), Vol.5, No.6, December 2013, pp:45-52. Airccse.org/journal/ijma.html

[26] Jain A, Ross A & Prabhakar S (2004) "An introduction to biometric recognition", IEEE Transactions on Circuits and Systems for Video Technology, Vol.14, No.1, pp4–20, Jan.

[27] Henry Rowley, Baluja S. & Kanade T. (1999) "Neural Network-Based Face Detection, Computer Vision and Pattern Recognition", Neural Network-Based Face Detection, Pitts-burgh, Carnegie Mellon University, PhD thesis.

[28] KahKay Sung & Tomaso Poggio (1994) Example Based Learning For View Based Human Face Detection, Massachusetts Institute of Technology Artificial Intelligence Laboratory and Center For Biological And Computational Learning, Memo 1521, CBCL Paper 112, MIT, December.

[29] Henry A. Rowley, Shumeet Baluja &Takeo Kanade. (1997) Rotation Invariant Neural Network-Based Face Detection, December, CMU-CS-97-201

[30] Jeffrey S. Norris (1999) Face Detection and Recognition in Office Environments, thesis, Dept. of Electrical Eng. and CS, Master of Eng in Electrical Eng., Massachusetts Institute of Technology.

[31] Hazem M. El-Bakry (2002), Face Detection Using Neural Networks and Image Decomposition Lecture Notes in Computer Science Vol. 22, pp:205-215.

[32] Lin-Lin Huang, et al (2003) "Face detection from cluttered images using a polynomial neural network", Neurocomputing, Vol.51, pp197 – 211.

[33] Masakazu Matsugu (2003) "Subject independent facial expression recognition with robust face detection using a convolutional neural network", Neural Networks, Vol.16, pp555–559.

[34] Stefan W., Christian I. & Uwe H (2004) "Evolutionary Optimization of Neural Networks for Face Detection", Proceedings of the 12th European Symposium on Artificial Neural Networks, Evere, Belgium: d-side publications.

[35] Marian Beszedes & Milos Oravec (2005) "A System For Localization Of Human Faces In Images Using Neural Networks", Journal Of Electrical Engineering, Vol. 56, No 7-8, pp195–199.

[36] N. Shilbayeh and G. Al-Qudah (2008) "Face Detection System Based On MLP Neural Network", Recent Advances in Neural Networks, Fuzzy Systems & Evolutionary Computing, ISSN: 1790-5109, ISBN: 978-960-474-195-3, pp 238-243.

[37] Zoran Bojkovic & Andreja Samcovic (2006) Face Detection Approach In Neural Network Based Method For Video Surveillance, 8th Seminar on Neural Network Applications in Electrical Engineering, Neurel, Faculty Of Electrical Eng., University Of Belgrade, Serbia, September 25-27.

[38] Aamer Mohamed, et al (2008) "Face Detection based Neural Networks using Robust Skin Color Segmentation", 5th International Multi-Conference on Systems, Signals and Devices, IEEE.

[39] Sahoolizadeh, Sarikhanimoghadam and Dehghani (2008) "Face Detection using Gabor Wavelets and Neural Networks", World Academy of Science, Engineering and Technology, Vol. 45, pp552- 554.

[40] Avinash Kaushal, J P S Raina (2010) "Face Detection using Neural Network & Gabor Wavelet Transform", International Journal of Computer Science and Technology (IJCST), Vol. 1, Issue.1, pp58-63, September 2010, ISSN : 0976 - 8491.

[41] Mohammad Abadi, et al, (2011) "Face Detection with the Help of Gabor Wavelets Characteristics and Neural Network Classifier", American Journal of Scientific Research, Issue.36, pp67-76, ISSN 1450-223X, http://www.eurojournals.com/ajsr.htm

[42] Anissa Bouzalmat, et al (2011) "Face Detection And Recognition Using Back Propagation Neural Network And Fourier Gabor Filters", Signal & Image Processing: An International Journal (SIPIJ) Vol. 2, No. 3, September 2011, DOI : 10.5121/sipij.2011.2302 15

[43] Kalavdekar Prakash N. (2010) "Face Detection using Neural Network", International Journal of Computer Applications (0975 – 8887),Vol.1, No.14, pp36-39.

[44] Mansaf M Elmansori & Khairuddin Omar (2011) "An Enhanced Face Detection Method Using Skin Color and Back-Propagation Neural Network", European Journal of Scientific Research, ISSN 1450-216X, Vol.55 No.1, pp80-86, http://www.eurojournals.com/ejsr.html

[45] Zuo F & P.H.N. de With (2008) "Fast face detection using a cascade of neural network ensembles", EURASIP Journal on Advances in Signal Processing, Volume 2008, Article ID 736508, Hindawi Publishing Corporation, pp1-13, doi:10.1155/2008/736508

[46] Hudson, Hagan and Demuth, Neural Network Toolbox™ User's Guide R2012a, The MathWorks, Inc., 3 Apple Hill Drive Natick, MA 01760-2098, 2012,  www.mathworks.com

## AUTHOR

Dr. Omaima N. A. Al-Allaf received the M.Sc. degree in CS from the Dept. of CS\ Faculty of Computers and Mathematical Sciences\ University of Mosul\ Mosul\ Iraq in 1999. She received the Ph.D. degree in CIS from CIS department\ Faculty of IS and Technology\ AABFS \ Jordan in 2008. Currently, she is an Assistant Professor at CIS department\ Faculty of Sciences and IT\ AlZaytoonah University of Jordan\ Amman\ Jordan (from 2009). Her research interests include image compression and recognition, Artificial Neural Networks, Genetic Algorithms. She is member of International Association of Engineers (IAENG).

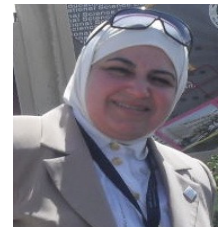